\documentclass{llncs}

\usepackage{makeidx}  

\usepackage{amsmath,algorithm,algorithmic,amsmath,color,comment,courier,dcolumn,float,graphicx,helvet,latexsym,subfigure,tikz,times,url}

\makeindex

\frontmatter          
\title{Topical Stance Detection for Twitter: A Two-Phase LSTM Model Using Attention}
\titlerunning{Topical Stance Detection for Twitter: A Two-Phase LSTM Model Using Attention}

\author{Kuntal Dey, Ritvik Shrivastava, Saroj Kaushik}
\institute{
Kuntal Dey, IBM Research, New Delhi, India. \email{kuntadey@in.ibm.com}
\and Ritvik Shrivastava, NSIT, New Delhi, India. \email{ritviks.it@nsit.net.in}
\and Saroj Kaushik, IIT, Delhi, India. \email{saroj@cse.iitd.ac.in}
}

\date{}

\begin{document}
\maketitle

\begin{abstract}
The topical stance detection problem addresses detecting the stance of the text content with respect to a given topic: whether the sentiment of the given text content is in \textsc{favor} of (positive), is \textsc{against} (negative), or is \textsc{none} (neutral) towards the given topic.
Using the concept of attention, we develop a two-phase solution.
In the first phase, we classify subjectivity - whether a given tweet is neutral or subjective with respect to the given topic.
In the second phase, we classify sentiment of the subjective tweets (ignoring the neutral tweets) - whether a given subjective tweet has a \textsc{favor} or \textsc{against} stance towards the topic.
We propose a Long Short-Term memory (LSTM) based deep neural network for each phase, and embed attention at each of the phases.
On the SemEval 2016 stance detection Twitter task dataset \cite{mohammad2016semeval}, we obtain a best-case macro F-score of 68.84\% and a best-case accuracy of 60.2\%, outperforming the existing deep learning based solutions.
Our framework, \textsc{T-PAN}, is the first in the topical stance detection literature, that uses deep learning within a two-phase architecture.
\end{abstract}

\section{Introduction}
\label{sec:intro}

Twitter, a hotbed of user generated content, has recently found traction among the researchers for the problem of topical stance detection.
Topical stance detection is the problem of finding whether a given tweet takes a \textsc{favor} (positive), \textsc{against} (negative) or \textsc{none} (neutral) stance towards a given topic.
It is at core of the opinion polarity detection and mining problem.
The problem is useful to solve in several practical scenarios, such as detecting user stance towards aspects of political, economic and social events, understanding stance-specific information propagation behavior of users {\it etc}.

\subsection{Related Work}
\label{subsec:relwork}

Sentiment detection from user-generated content has been a long-standing problem \cite{rosenthal2014semeval}.
However, stance detection, where the sentiment (opinion) of the user is not generic but with respect to a specific topic, has gained research attention only in recent times.
A seminal work by Mohammad {\it et al.} \cite{mohammad2016stance}, followed by a SemEval 2016 task \cite{mohammad2016semeval} conducted by the authors, resulted in starting wide-spread research in the area.

Different models, including traditional machine learning approaches, genetic algorithms, and deep learning approaches such as convolutional neural networks (CNN), recurrent neural networks (RNN) and long short-term memory (LSTM), were proposed in the SemEval 2016 topical stance detection contest.
MITRE \cite{zarrella2016mitre} provided the best deep learning solution in the contest, initializing weights from a 256-dimensional word embeddings learned using the word2vec skip-gram algorithm \cite{mikolov2013efficient}, followed by a second layer with 128 LSTM units.
Among others, {\it pkudblab} \cite{wei2016pkudblab} and DeepStance \cite{vijayaraghavan2016deepstance} use deep CNN models.
Augenstein {\it et al.} \cite{augenstein2016stance} employ a bidirectional attention model.

Some works used a two-phase approach.
ECNU \cite{zhang2016ecnu}, in the first phase, determines whether a given tweet is relevant to a given target topic, and in the second phase, detects orientation (favor/against).
The work by ltl.uni-due \cite{wojatzki2016ltl} also uses a two-level stacked classifier approach using Support Vector Machines (SVM).
Among others, TakeLab \cite{boltuzic2016takelab}, mixed machine learning with genetic algorithms.
Other approaches, such as CU-GWU \cite{elfardy2016cu} and IUCL-RF \cite{liu2016iucl}, employed traditional machine learning.
A shared task has also been proposed recently \cite{taule2017overview}.

The overall average values of F-scores, obtained by the task participants, ranged from $46.19$ at the lower end to all the way up to $67.82$ at the higher end.
A recent work was conducted by Du {\it et al.} \cite{du2017stance}, the first of its kind that deeply ingrained the stance words in the architecture and used attention modeling.
It outperformed the deep learning based approaches, attaining F-score of 68.79\% as against the deep-learning state of the art F-score of 67.82\%.
We further observe that, the SemEval 2016 tasks were evaluated as a macro average of the F-score for only the {\it favor} and {\it against}, ignoring the {\it none} (neutral) class.
We, however, perform accuracy measurements against all the three classes as well (in addition to the F-score that we measure following the traditional literature), and show that our model outperforms the best-known deep learning system not only for two-class macro average F-score, but for a full three-class accuracy measure as well.

\subsection{Our Contributions}
\label{subsec:contributions}

We propose a two-phase approach, using attention embedding at each phase and encoding using LSTM.
The given SemEval 2016 \cite{mohammad2016semeval} dataset contains three classes - \textsc{favor}, \textsc{against} and \textsc{none}.
Our work is based on the observation that messages with neutral stances are usually non-subjective, while the ones with favor and against stances are usually subjective.
Thus, in the first phase of our two-phase approach, we use a LSTM to detect subjectivity, and classify into subjective (non-neutral) versus neutral (none).
And in the second phase, we use another LSTM to detect sentiment (favor/against) of the tweets that were labeled subjective in the earlier phase.
Akin to the philosophy of Du {\it et al.} \cite{du2017stance}, we also use an attention model, and deeply embed the topical attention as part of the input to the classifier.
Since a given tweet does not necessarily contain the topic against which the stance is sought for, this step plays an important role in transforming the learning into a topic-specific learning.
This is absent in the literature except for Du {\it et al.} \cite{du2017stance}.
Our model thus is the first of its kind, that uses a two-phase LSTM-based architecture with attention embedding ingrained.

The contributions of our work are the following.
\begin{itemize}
\item We propose \textsc{T-PAN}, a two-phase attention-embedded LSTM-based approach for detecting stance of tweets towards given topics.
\item In the first phase, we perform subjectivity analysis of the tweets, using a combination of LSTM and attention embedding.
\item In the second phase, we perform sentiment analysis on the subjective tweets, again using a combination of LSTM and attention embedding.
\item Empirically, on the SemEval 2016 benchmark dataset, we demonstrate the effectiveness of our system. Our model is novel, and we outperform the deep learning based literature in terms of accuracy (60.2\% against 58.7\%), as well as F-score (68.84\% against 68.79\%).
\end{itemize}

\section{Central Idea}
\label{sec:algo}

\subsection{Approach Overview}
\label{subsec:overview}

\begin{figure*}[htb]
\centering
\includegraphics[width=0.9\textwidth]{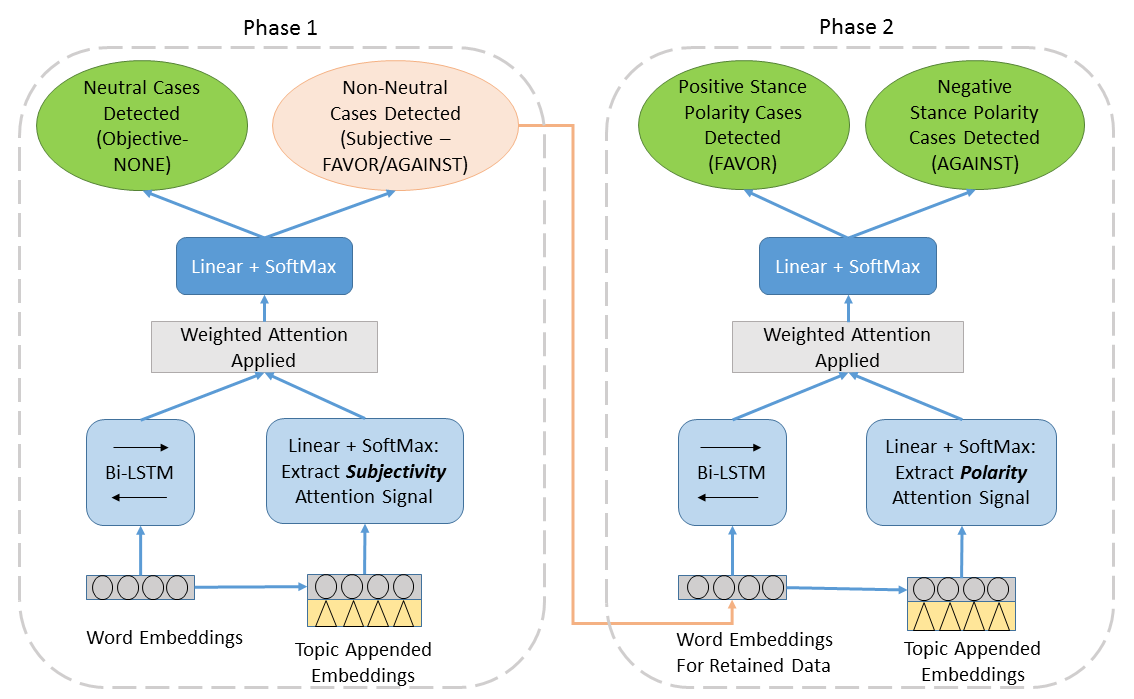}
\caption{System Architecture Diagram}
\label{fig:arch}
\vspace{-0.2in}
\end{figure*}

Table~\ref{tab:examples} shows a few randomly chosen samples from the training set across topics, to provide the reader with an intuition of the data available.
As mentioned earlier, our task comprises of three classes of data: \textsc{favor}, \textsc{against} and \textsc{none}.
While {\it favor} and {\it against} tweets are often subjective in nature, the {\it neutral} tweets often are non-subjective.
The architecture of our system is presented on Figure~\ref{fig:arch}.

Our model is a two-phase one.
At each phase, there are two components - a bi-directional LSTM and an attention mechanism.
The bi-directional LSTM is used for feature encoding.
The attention logic uses augmentation of the word embeddings with target topics, and subsequently passes it through a linear layer for computing attention of each word in the text in the context of the topic under consideration.

\begin{table*}[tbh]
\begin{center}
\vspace{-0.15in}
\scriptsize
\begin{tabular}{|l|p{8cm}|l|}
\hline
\textbf{Target} & \textbf{Tweet} & \textbf{Stance} \\
\hline
\multicolumn{3}{|l|}{\textit{Examples from the {\bf favor} stance}} \\
\hline
Atheism & Everyone is able to believe in whatever they want. \#Freedom & FAVOR \\
Feminist Movement & @OliviaJeniferx it's not always the guys job. \#equality & FAVOR \\
\hline
\multicolumn{3}{|l|}{\textit{Examples from the {\bf against} stance}} \\
\hline
Atheism & Be still. Be patient. Watch and let God work. & AGAINST \\
Feminist Movement & Friendly reminder that the ``Gender Pay Gap'' is a myth. & AGAINST \\
\hline
\multicolumn{3}{|l|}{\textit{Examples from the {\bf none} stance}} \\
\hline
Atheism & Alot of angry people in this world. Peace to all. \#love & NONE \\
Feminist Movement & @sass\_unicorn lol! Young male children for & NONE \\
\hline
\end{tabular}
\end{center}
\caption{Random examples of tweets of the different stances, for a few of the given target topics}
\label{tab:examples}
\vspace{-0.5in}
\end{table*}

\subsection{Embedding Augmentation with Target Topics and Determining Attention}
\label{subsec:augmentation}

To compute attention, we augment the embedding of the constituent words with the average embedding of the target.
If the words in a given target topic comprises of word embeddings $\{\tilde{z_1}, \tilde{z_2}, ..., \tilde{z_n}\}$, then we compute the embedding of the target topic $\tilde{z}$ as $\tilde{z} = \frac{\sum\limits_{i=1}^n \tilde{z_i}}{|n|}$.
The words within the sentence, that have the embeddings $\{z_1, z_2, ..., z_m\}$ of dimension $d_z$, are thus augmented with dimension $d_{\tilde{z}}$ (the dimension of $\tilde{z}$), and each word gets a new embedding dimension of $d_{z}+d_{\tilde{z}}$.
This is processed as depicted in Figure~\ref{fig:arch}, by first passing via a linear layer followed by a softmax, and subsequently ingraining the attention derived for each word into the LSTM-encoded features, using a product of the LSTM-embedded features and the output of the linear layer.
We note that, while our approach is largely different from Du {\it et al.} \cite{du2017stance} in terms of the overall system architecture (our approach is two-phase while theirs is one-phase), the philosophy of augmenting each word of the sentence with the average embedding of the target topic words is similar.

\subsection{Training the Models}
\label{subsec:training}

Using a similar underlying architecture, the first phase is trained for subjectivity, and the second phase for sentiment polarity.
Hence, the attention gets trained for subjectivity in the first phase and for polarity in the second phase.
The subjective outputs of the first phase are passed through the second phase, while the rest (non-subjective) are assigned a class label \textsc{none} and kept aside.
We try using both SGD (stochastic gradient descent) as well as Adam optimizers for experiments, and these yield similar effectiveness.
We train our model using cross-entropy loss function.
The loss of one phase is not propagated to the other.

\section{Experiments}
\label{sec:expt}

\subsection{Data Description}
\label{subsec:datadesc}

We use the benchmark training and test data provided by the SemEval 2016 stance detection task \cite{mohammad2016semeval}.
For self-containment, we reproduce their data, in Table~\ref{tab:datasesc}.
We use the evaluation script they provide, for calculating F-score.
Further, since their script only accounts for the \textsc{favor} and \textsc{against} classes and computes a macro F-score as average of the two (ignoring the \textsc{none} class), we develop an additional script to calculate the accuracy using the three classes, as a ratio to the total number of correct predictions to the total test data size.
We use PyTorch for programming.
We perform data cleaning: net slang removal (for tweet normalization) using an online dictionary\footnote{http://www.noslang.com/dictionary} and stopword removal using a Stanford NLP resource for stopword removal\footnote{https://nlp.stanford.edu/IR-book/html/htmledition/dropping-common-terms-stop-words-1.html}.

\begin{table*}[tbh]
\begin{center}
\scriptsize
\begin{tabular}{|l|c|c|c c c|c|c c c|}
\hline
& & & \multicolumn{3}{c|}{\textbf{\% of instances in Train}} & & \multicolumn{3}{c|}{\textbf{\% of instances in Test}} \\
\textbf{Target} & \textbf{\#total} & \textbf{\#train} & favor & against & neither & \textbf{\#test} & favor & against & neither \\
\hline
Atheism & 733 & 513 & 17.9 & 59.3 & 22.8 & 220 & 14.5 & 72.7 & 12.7 \\
C.C.C. & 564 & 395 & 53.7 & 3.8 & 42.5 & 169 & 72.8 & 6.5 & 20.7 \\
Feminist Movement & 949 & 664 & 31.6 & 49.4 & 19.0 & 285 & 20.4 & 64.2 & 15.4 \\
Hillary Clinton & 984 & 689 & 17.1 & 57.0 & 25.8 & 295 & 15.3 & 58.3 & 26.4 \\
L.A. & 933 & 653 & 18.5 & 54.4 & 27.1 & 280 & 16.4 & 67.5 & 16.1 \\ \hline
All & 4,163 & 2,914 & 25.8 & 47.9 & 26.3 & 1,249 & 24.3 & 57.3 & 18.4 \\
\hline
\end{tabular}
\end{center}
\caption{Data for the SemEval 2016 stance detection task. Target C.C.C. $\rightarrow$ Climate Change is Concern. Target L.A. $\rightarrow$ Legalization of Abortion. Table courtesy: \cite{mohammad2016semeval}.}
\label{tab:datasesc}
\vspace{-0.3in}
\end{table*}

\subsection{Performance of Our Model T-PAN and Its Constituent Components}
\label{subsec:performance}

Our system delivers commendable performance for detecting the user stances towards the individual topics, as well as, a robust overall performance across the topics.
We empirically observe the performance of our T-PAN model.
We further examine the performance of different LSTM-based architectures that eventually are composed to develop our end-to-end framework.
Table~\ref{tab:performance} provides the details of the performance attained by the full T-PAN model, as well as the impact of performance of the constituent LSTM blocks and configurations by systematic component ablation.

\begin{table*}[tbh]
\begin{center}
\scriptsize
\begin{tabular}{|l|l|c|}
\hline
\textbf{Phase 1} & \textbf{Phase 2} & \textbf{Accuracy} \\
\hline
Bi-LSTM & Bi-LSTM & 57.08  \\
Bi-LSTM + Tweet Cleaning & Bi-LSTM + Tweet Cleaning & 57.61  \\
Bi-LSTM & One-Phase Attention & 59.32 \\
Bi-LSTM & One-Phase Attention + Tweet Cleaning & 57.53 \\
Bi-LSTM + Tweet Cleaning & One-Phase Attention + Tweet Cleaning & 59.85 \\
One-Phase Attention & One-Phase Attention & \textbf{60.22} \\
One-Phase Attention + Tweet Cleaning & One-Phase Attention + Tweet Cleaning & \textbf{60.24} (T-PAN) \\
\hline
\multicolumn{2}{|c|}{Our implementation of TAN \cite{du2017stance}} & 58.76\\
\hline
\end{tabular}
\end{center}
\caption{Performance of the different underlying two-phase architectures.}
\label{tab:performance}
\vspace{-0.4in}
\end{table*}

\subsection{Comparing Our System Against the Deep-Learning Literature}
\label{subsec:ablation}

As observed in Table~\ref{tab:comparison}, our best system (the T-PAN model) outperforms the state of the art that uses deep neural networks for topical stance classification.
Out of the five given classes, we perform the best in one class, the {\it TAN} model \cite{du2017stance} outperforms us in two classes and the SemEval tasks perform better than our model (as well as better than the TAN model \cite{du2017stance}) for the other two classes.

\begin{table*}[tbh]
\begin{center}
\scriptsize
\begin{tabular}{|l|c c c c c |c|}
\hline
\textbf{Target} & \textbf{NBOW} & \textbf{LSTM} & \textbf{LSTM}$_E$ & \textbf{TOP Sem-Eval} & \textbf{TAN} & \textbf{\textit{T-PAN}} \\
\hline
Atheism & 55.12 & 58.18 & 59.77 & \textbf{61.47} & 59.33 & 61.19  \\
C.C.C. & 39.93 & 40.05 & 48.98 & 41.63 & 53.59 & \textbf{66.27}  \\
Feminist Movement & 50.21 & 49.06 & 52.04 & \textbf{62.09} & 55.77 & 58.45 \\
Hillary Clinton & 55.98 & 61.84 & 56.89 & 57.67 & \textbf{65.38} & 57.48  \\
L.A. & 55.07 & 51.03 & 60.34 & 57.28 & \textbf{63.72} & 60.21  \\ \hline
Overall & 60.19 & 63.21 & 66.24 & 67.82 & 68.79 & \textbf{68.84} \\
\hline
\end{tabular}
\end{center}
\caption{Comparing F-scores of different models. A part of the table has been replicated from Du {\it et al.} \cite{du2017stance}. NBOW $\gets$ Neural Bag-of-Words. LSTM $\gets$ LSTM without target-specific embedding. LSTM$_E$ $\gets$ LSTM with target-specific embedding, by \cite{du2017stance}. TOP Sem-Eval $\gets$ The best-reported systems in SemEval 2016. TAN $\gets$ The final output of \cite{du2017stance}. T-PAN $\gets$ Our framework.}
\label{tab:comparison}
\vspace{-0.3in}
\end{table*}

\section{Conclusion}
\label{sec:concl}

We proposed T-PAN, a two-phase LSTM-based model with attention embedding, for detecting user stance with respect to given topics on Twitter.
First, we classified the tweets into two: neutral and non-neutral, where non-neutral comprised of favor and against stances.
Second, we classified the tweets labeled as non-neutral in the first phase, into two - favor and against stances.
In each phase, we encoded the input sentences in form of a sequence of words using a bi-directional LSTM, and attention embedding.
We investigated the impact of embedding topical attention, as well as, the impact of different LSTM architectures, on our approach.
We empirically demonstrated the robustness of our framework T-PAN, by delivering the highest-known performance among all the deep learning approaches.
Our model is easy to implement, reusable and practicable.

\bibliographystyle{splncs03}
\bibliography{bib}

\end{document}